\documentclass[fleqn,10pt]{olplainarticle}

\usepackage{algorithm} 
\usepackage{subfigure}
\usepackage{multicol}
\usepackage{algpseudocode} 
\algnewcommand{\LeftComment}[1]{\Statex \(\triangleright\) #1}
\usepackage{subfigure}
\graphicspath{{.}}

\newcommand{\x}{\mathbf{x}}
\newcommand{\rt}{\mathbf{r}}
\newcommand{\ctr}{\bar{\mathbf{x}}}
\newcommand{\X}{\mathbf{X}}
\newcommand{\G}{\mathbf{G}}

\newcommand{\cd}{\kappa}
\DeclareMathAlphabet{\mathcal}{OMS}{cmsy}{m}{n}
\SetMathAlphabet{\mathcal}{bold}{OMS}{cmsy}{b}{n}

\newcommand\footer[1]{%
  \begingroup
  \renewcommand\thefootnote{}\footnote{#1}%
  \addtocounter{footnote}{-1}%
  \endgroup
}

\title{FLASC: A Flare-Sensitive Clustering Algorithm}

\author[1]{Dani\"el M. Bot}
\author[1]{Jannes Peeters}
\author[2]{Jori Liesenborgs}
\author[3]{Jan Aerts}
\affil[1]{Data Science Institute (DSI), UHasselt, Diepenbeek, Belgium}
\affil[2]{Expertisecentrum voor Digitale Media (EDM), UHasselt -- Flanders Make, Diepenbeek, Belgium}
\affil[3]{Augmented Intelligence for Data Analytics (AIDA) Lab, Department of Biosystems, KU Leuven, Leuven, Belgium}

\begin{abstract}
  Clustering algorithms are often used to find subpopulations in exploratory
  data analysis workflows. Not only the clusters themselves, but also their
  shape can represent meaningful subpopulations. In this paper, we present
  FLASC, an algorithm that detects branches within clusters to identify such
  subpopulations. FLASC builds upon HDBSCAN*---a state-of-the-art density-based
  clustering algorithm---and detects branches in a post-processing step that
  describes within-cluster connectivity. Two variants of the algorithm are
  presented, which trade computational cost for noise robustness. We show that
  both variants scale similarly to HDBSCAN* in terms of computational cost and
  provide stable outputs using synthetic data sets, resulting in an efficient
  flare-sensitive clustering algorithm. In addition, we demonstrate the benefit
  of branch-detection on two real-world data sets.
\end{abstract}


\begin{document}
\flushbottom
\maketitle
\thispagestyle{empty}
\footer{This work is licensed under a Creative Commons
  Attribution-NonCommercial-NoDerivatives 4.0 International License. This work
  has been submitted to PeerJ for possible publication. Copyright may be
  transferred without notice, after which this version may no longer be
  accessible.}

\begin{multicols}{2}
\section*{Introduction}%
\label{sec:introduction}
Exploratory Data Analysis (EDA)---i.e., searching for interesting patterns in
data---is ubiquitous in data science and knowledge discovery workflows.
Detecting which subpopulations exist in a data set is a common step in EDA.
Typically, subpopulations are detected as clusters, which traditional algorithms
model as Gaussian distributions (as cited in~\cite{campello2015hdbscan}). More
recently, density-based clustering has become more popular
(e.g.,~\cite{ester1996dbscan,campello2013density}). These algorithms,
informally, specify clusters as regions of high density separated by regions of
lower density, allowing them to capture cluster shapes, which may be non-convex
and reveal relevant subpopulations. For example, a Y-shaped cluster might
represent an evolving process with two distinct outcomes. Consequently, the
branches in a cluster's manifold---i.e., \emph{flares}---can represent
meaningful subpopulations (see also,
\cite{lum2013mapper,skaf2022review,kamruzzaman2018detecting,
reaven1979diabetes}). 

Clustering algorithms generally cannot detect this type of subgroup because
there is no gap that separates flares from their cluster. From a topological
perspective, clustering algorithms describe the connected components in a
\emph{simplicial complex} of the data~\citep{carlsson2014flares}: a set of
points, edges, and triangles that describe connectivity. Flares are connected in
the simplicial complex. In other words, there is a path between data points in
different branches that exclusively goes through data points `that lie close
together'. Therefore, they have a vanishing homology and cannot be detected as
clusters.

Several flare-detection techniques have been proposed in topological data
analysis literature. For example, \cite{carlsson2014flares} proposed
\emph{functional persistence} to distinguish flares from a data set's central
core. This technique quantifies data point centrality as the sum of its
distances. Central observations have lower distance sums than points towards the
extreme ends of the feature space. A manually controlled centrality threshold
then removes the data's core, separating branches from each other and making
them detectable as clusters.

Extending this approach to compute branching hierarchies requires considering
both the centrality and the data point distances in what is called a
\emph{bi-filtration}. The centrality controls how much of the core is retained
to describe how branches grow and merge. The data point distances determine
whether points are connected and form a cluster. Algorithms for computing
bi-filtrations are computationally
expensive~\citep{lesnick2019minpres,kerber2021minpres}. Their resulting
\emph{bi-graded} hierarchies are also complicated to work with, as they do not
have a compact representation~\citep{carlsson2014flares} (though research into
usable representations is ongoing~\citep{botnan2022multipersistence}), and
existing visualisations are
non-trivial~\citep{lesnick2015rivet,scoccola2023persistable}. Alternative
strategies that simultaneously vary both dimensions in a single-parameter
filtration exist~\citep{chazal2009gromov}; however, they remain computationally
expensive~\citep{vandaele2021stable}.

In the present paper, we present an approach that efficiently computes branching
hierarchies and detects branch-based subgroups of clusters in unfamiliar data.
Inspired by~\cite{vandaele2021stable}, we compute branching hierarchies using
graph approximations of the data. This effectively replaces functional
persistence's manual centrality threshold with the question of which data points
should be connected in the approximation graph. We will use
HDBSCAN*~\citep{campello2013density,campello2015hdbscan}---a state-of-the-art
density-based clustering algorithm---to answer this question. Conceptually, our
approach can be thought of as creating a sequence of subgraphs which
progressively include more and more central points and tracking the remaining
connected components. Interestingly, a similar method has been used by
\cite{li2017branches} to detect \emph{actual} branches in 3D models of plants.

Our main contribution is this flare detection approach, implemented as a
post-processing step in \cite{mcinnes2017hdbscan}'s HDBSCAN*
implementation\footnote{\texttt{hdbscan}:
\url{https://github.com/scikit-learn-contrib/hdbscan}} and as a stand-alone
package\footnote{\texttt{pyflasc}: \url{https://github.com/vda-lab/pyflasc}}. We
propose two types of approximation graphs that naturally arise from HDBSCAN*'s
design and provide a practical centrality metric for points within a cluster
that is computable with a linear complexity. Combining density-based clustering
and flare detection into a single algorithm provides several attractive
properties:
\begin{itemize}
  \item The ability to detect clusters and their branches.
  \item No manual thresholds, instead intuitive minimum cluster and branch sizes
  control the detected subgroups.
  \item Low computational cost compared to multi-parameter persistence and other
  structure learning algorithms.
  \item High branch-detection sensitivity and noise robustness by operating on
  HDBSCAN*-clusters, which suppress spurious noisy connectivity. 
  \item Branch-detection at multiple distance scales because each cluster has
  its own approximation graph.
\end{itemize}
We call the resulting algorithm flare-sensitive clustering (FLASC) and
empirically analyse its computational cost and stability on synthetic data sets
to show that the flare detection cost is relatively low. In addition, we
demonstrate FLASC on two real-world data sets, illustrating its benefits for
data exploration.

The remainder of this article is organised as follows: Section `Related Work'
provides a literature overview of related data analysis algorithms and describes
the HDBSCAN* algorithm in more detail. Section `The FLASC Algorithm' describes
how FLASC builds on HDBSCAN* to detect branches within clusters and discusses
the algorithm's complexity and stability. Section `Experiments' presents our
empirical analyses that demonstrate the algorithm's computational complexity,
stability, and benefits for data exploration. Finally, Sections `Discussion' and
`Conclusion' discuss our results and present our conclusions.

\section*{Related Work}%
\label{sec:related}
The purpose of our work is detecting branching structures within clusters. As
such, our work relates to manifold and structure learning algorithms in general.
In this section, we provide an overview of related data analysis algorithms and
introduce HDBSCAN*, the density-based clustering algorithm we build upon.

\subsection*{Structure Learning}%
\label{sec:related:structure-learning}
Many types of data lie not just on a manifold, but on a smooth, one-dimensional
structure. Extracting these structures can be essential in unsupervised learning
applications. For example, road networks can be extracted from GPS
measurements~\citep{bonnaire2022mixturestructure} and cell developmental
trajectories can be extracted from gene expression
data~\citep{qiu2017graphembedding,vandaele2020boundary}. Algorithms for
extracting such structures are related to our work because the branch-based
subgroups we are interested in can be extracted from them by partitioning the
data between their intersections~\citep{chervov2020graphquality}.

Most work on extracting smooth, one-dimensional manifolds is based on
\cite{hastie1989principalcurve}'s concept of \emph{principal curves}: a smooth,
self-consistent curve that passes through the middle of the data. Techniques
estimating principal curves, trees, or graphs are often based on Expectation
Maximisation~\citep{dempster1977em} and optimise the one-dimensional manifold
directly (e.g.,~\cite{bonnaire2022mixturestructure,mao2017graphembedding}).
Alternative approaches are more closely related to non-linear Dimensionality
Reduction (DR) algorithms that model the data's structure as an undirected graph
(e.g.,~\cite{roweis2000lle,
tenenbaum2000isomap,belkin2003laplacianeigenmaps,vandermaaten2008tsne,mcinnes2018umap}).
For example, \cite{vandaele2020boundary} use (manually) pruned minimum spanning
trees over edges weighted by their boundary coefficient to extract a graph's
backbone. Alternatively, \cite{ge2011reebskeleton} extract graph skeletons using
a \emph{Reeb Graph}. Reeb Graphs track the existence, merges, and splits of
connected components in level sets of a continuous function defined on a
manifold. Using geodesic distances to an arbitrary eccentric point as the
continuous function makes the Reeb Graph capture the manifold's skeleton.
Interestingly, Mapper---an algorithm that approximates Reeb
Graphs~\citep{singh2007mapper}---has also been used for detecting branch-based
subpopulations~\citep{kamruzzaman2018detecting}.

There are several similarities between these methods and our work. Like
\cite{vandaele2020boundary}, our approach detects tree-based branching
hierarchies. Like \cite{ge2011reebskeleton}, our approach is topologically
inspired. Where they create a Reeb Graph, we compute a Join Tree. The main
difference of these methods compared to our work is their goal. We aim to
identify relevant branch-based subpopulations. \cite{ge2011reebskeleton},
\cite{vandaele2020boundary}, and the expectation maximisation-based algorithms
explicitly model the data's structure which necessitates a larger computational
cost. 

\subsection*{HDBSCAN*}%
\label{sec:related:hdbscan}
\begin{figure*}[t!]
  \centering
  \subfigure[2D data]{\includegraphics[width=.245\textwidth]{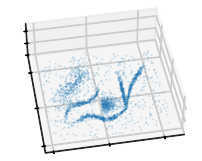}\label{fig:density-clustering:data}}%
  \subfigure[Density profile]{\includegraphics[width=.245\textwidth]{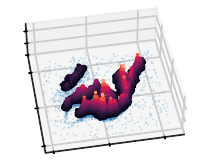}\label{fig:density-clustering:profile}}%
  \subfigure[HDBSCAN* clusters]{\includegraphics[width=.245\textwidth]{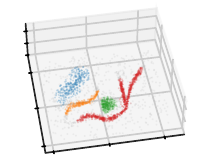}\label{fig:density-clustering:clusters}}%
  \subfigure[Cluster hierarchy]{\includegraphics[width=.245\textwidth]{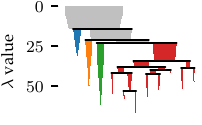}\label{fig:density-clustering:hierarchy}}%
  \caption{Density-based clustering concepts behind HDBSCAN*. (a) A 2D example
  point cloud with varying density adapted
  from~\cite{mcinnes2022documentation}'s online tutorial. (b) Density contours
  in a height map illustrate the data's density profile. Peaks in this density
  profile correspond to \emph{density contour clusters}. (c) Clusters extracted
  from the density profile by HDBSCAN* indicated in colour. (d) The
  \emph{density contour tree} describes how density contour clusters merge when
  considering lower density thresholds.}%
  \label{fig:density-clustering}
\end{figure*}
HDBSCAN* is a density-based clustering algorithm that provides state-of-the-art
clustering performance~\citep{campello2013density,campello2015hdbscan}.
Informally, density-based clustering specifies clusters as regions of high
density separated by regions of lower density. This formulation does not limit
clusters to convex shapes and provides a natural way to separate noise points
from clusters. The algorithm is well suited for exploring unfamiliar data
because---unlike older popular clustering algorithms---HDBSCAN* does not require
the number of clusters or the distance between clusters to be specified in
advance.

Several studies have implemented and adapted the HDBSCAN* algorithm:
\cite{mcinnes2017accelerated} improved the algorithm's computational performance
by using \emph{space trees} for finding the data points' nearest neighbours and
provide a popular and efficient Python
implementation~\citep{mcinnes2017hdbscan}. \cite{stewart2022java} created a Java
implementation with a novel prediction technique for unseen data points.
\cite{jackson2018knngraph} presented an approximate HDBSCAN* algorithm that uses
NN-descent~\citep{dong2011nndescent} for finding the nearest neighbours
providing fast distributed performance. \cite{malzer2020hybrid} introduced a
cluster selection distance threshold that effectively creates a hybrid between
DBSCAN's~\citep{ester1996dbscan} and HDBSCAN*'s cluster selection, improving the
algorithm's performance on data sets with small clusters and a large density
variability. \cite{neto2021multiscale} showed how Relative Neighbourhood Graphs
(RNGs)~\citep{toussaint1980rng} can be used to efficiently compute HDBSCAN*
cluster hierarchies for multiple \emph{min cluster size} values. Their follow-up
work presented MustaCHE, a visualisation tool for the resulting meta-cluster
hierarchy~\citep{neto2018mustache}. To our knowledge, no previous study has
adapted HDBSCAN* for detecting flares.

Because our work builds on HDBSCAN*, it is relevant to explain how the algorithm
works in more detail. The remainder of this section describes HDBSCAN* following
\cite{campello2015hdbscan}'s explanation. We refer the reader to
\cite{mcinnes2017accelerated} for a more formal, statistically motivated
description of the algorithm.

\subsubsection*{The HDBSCAN* Algorithm}%
\label{sec:related:hdbscan:algorithm}
\begin{figure*}[t!]
  \centering
  \subfigure[Cluster centroid]{\includegraphics[width=.245\textwidth]{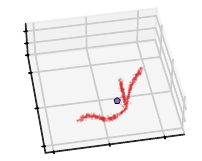}\label{fig:eccentricity-clustering:data}}%
  \subfigure[Eccentricity profile]{\includegraphics[width=.245\textwidth]{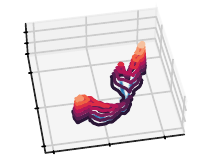}\label{fig:eccentricity-clustering:profile}}%
  \subfigure[FLASC branches]{\includegraphics[width=.245\textwidth]{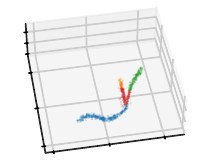}\label{fig:eccentricity-clustering:clusters}}%
  \subfigure[Branch hierarchy]{\includegraphics[width=.245\textwidth]{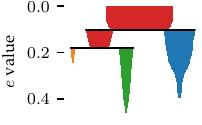}\label{fig:eccentricity-clustering:hierarchy}}%
  \caption{Density-based clustering concepts behind FLASC. (a) A within-cluster
  eccentricity $e(\x_i)$ is defined for each point $\x_i$ in cluster $C_j$ based
  on distances to the cluster's membership weighted average shown by the
  pentagon mark. (b) The cluster's eccentricity profile visualised as contours
  on a height map. Peaks in the profile correspond to branches in the cluster.
  (c) Branches extracted from the cluster by FLASC indicated in colour. The
  cluster's centre is given its own label. (d) The \emph{eccentricity contour
  tree} describes how branches merge when considering lower eccentricity
  thresholds.}%
  \label{fig:eccentricity-clustering}
\end{figure*}
HDBSCAN* is based on density-based clustering concepts pioneered
by~\cite{wishart1969mode} and formalised by~\cite{hartigan1975clustering}. We
demonstrate these ideas using a 2D point cloud adapted from
\cite{mcinnes2022documentation}'s online tutorial shown in
Fig.~\ref{fig:density-clustering:data}. In general, let $\X = \{\x_1, \dots,
\x_N\}$ be a data set consisting of $N$ feature vectors $\x_{(\cdot)}$ and a
distance metric $d(\x_i, \x_j)$. Then, a point's \emph{core distance}
$\cd(\x_i)$ is the distance to its $k$-nearest neighbour and HDBSCAN* estimates
its density as $\lambda_k(\x_i) = 1 / \cd(\x_i)$~\citep{campello2015hdbscan}.
Figure~\ref{fig:density-clustering:profile} illustrates the example's density
profile as contours in a height map. \emph{Density contour clusters} intuitively
correspond to peaks in the density profile, for example, the clusters indicated
in colour in Fig.~\ref{fig:density-clustering:clusters}. More formally, the
density contour clusters at some threshold $\lambda_t$ are a collection of
maximal, connected subsets in a level set $\{\x\;|\;\lambda(\x) \ge
\lambda_t\}$~\citep{hartigan1975clustering}. In other words, density contour
clusters are the connected components of points with a density larger than some
threshold. \emph{Density contour trees} capture the hierarchy in which density
contour clusters merge as the density threshold decreases. From a topological
perspective, density contour trees are a \emph{join tree} of the data's density
profile.

Data sets generally do not have an inherent notion of connectivity between their
data points. Such a notion is needed to determine whether two points are part of
the same density contour cluster at a threshold $\lambda_t$. HDBSCAN* solves
this problem by considering points to be connected if the distance between them
is smaller than or equal to $1 / \lambda_t$. This solution is possible because
density is defined in terms of distance. HDBSCAN* uses a \emph{mutual
reachability} distance between points for this purpose, which is defined
as~\citep{campello2015hdbscan}:
\begin{equation}%
  \label{eq:mutual-reachability}
  d_{mreach}(\x_i, \x_j) = \begin{cases}
    \max{\left\{\cd(\x_i), \cd(\x_j), d(\x_i, \x_j)\right\}} & \text{if } \x_i \ne \x_j\text{,} \\
    0                                                        & \text{otherwise,}
  \end{cases}
\end{equation}
where the value of $k$ as used in $\cd$ is specified manually and acts as a
smoothing factor for the density estimation.

Using the mutual reachability distance to provide connectivity, we can now
recover the density contour tree. The edges that change connectivity between
density contour clusters are exactly those edges in the data's Minimum Spanning
Tree (MST) ~(see, f.i.,~\cite{doraiswamy2021topomap}). HDBSCAN* uses a MST to
efficiently compute a single linkage clustering
hierarchy~\citep{sibson1973slink}. The resulting dendrogram is simplified using
a manually specified minimum cluster size $m_c$ to recover a condensed cluster
hierarchy that resembles the data's density profile as shown in
Fig.~\ref{fig:density-clustering:hierarchy}. From the root down, only the sides
of a split containing more than $m_c$ points are considered to represent
clusters. Sides with fewer points are interpreted as ``falling out of the parent
cluster''~\citep{mcinnes2017accelerated} or the cluster disappearing completely.

HDBSCAN* provides two strategies for selecting clusters from the condensed
hierarchy: the \emph{excess of mass} (EOM) strategy and the \emph{leaf}
strategy~\citep{campello2015hdbscan}. The leaf strategy selects all leaf
segments in the condensed hierarchy, typically resulting in multiple small
clusters. The EOM strategy maximises relative cluster stability while preventing
any data point from being a member of more than one selected cluster. A cluster
$C_j$'s relative stability  $\sigma_k(C_j)$ is defined
as~\citep{campello2015hdbscan}: 
\begin{equation}%
  \label{eq:stability}
  \sigma_k(C_j) = \sum_{\x_i \in C_j}{\lambda_{k,max}^{C_j}(\x_i) - \lambda_{k,min}^{C_j}},
\end{equation}
where $\lambda_{k,max}^{C_j}(\x_i)$ is the density at which $\x_i$ falls out of
$C_j$ or $C_j$ separates into two clusters, and $\lambda_{k,min}^{C_i}$ is the
minimum density at which $C_j$ exists. In words, the stability of a cluster is
the sum of density ranges in which points are part of the cluster, which
corresponds to the area of the cluster's icicle in
Fig.~\ref{fig:density-clustering:hierarchy}. HDBSCAN*'s \emph{cluster selection
epsilon} parameter can be used to specify a minimum persistence for EOM
clusters~\citep{malzer2020hybrid}.

\section*{Flare-sensitive HDBSCAN*}%
\label{sec:fhdbscan}
\begin{algorithm*}[t!]
  \caption{A high-level overview of the FLASC algorithm.}%
  \label{alg:fhdbscan}
  \begin{algorithmic}[1]
    \Function {flasc}{$\X, d$} 
    \LeftComment{$\X$ is a dataset with $N$ feature vectors $\x_{(\cdot)}$ and
    $d$ is a distance metric $d(\x_i, \x_j)$.} 
    \State evaluate \Call{hdbscan}{$\X, d$} and store its internal data
    structures.
    \For {each detected cluster $C_j$} 
      \State compute the eccentricity $e(\x_i)$ for all $\x_i \in Cj$. 
      \State extract the \emph{cluster approximation graph} $\G_k^{C_j}$. 
      \State compute the single linkage clustering hierarchy of $\G_k^{C_j}$ 
      \State simplify the clustering hierarchy using a minimum branch size $m_b$. 
      \State extract labels and probabilities for a `flat' clustering.
    \EndFor 
    \State combine the cluster and branch labels and probabilities.
    \State \Return the membership labels and probabilities. 
    \EndFunction
  \end{algorithmic}%
\end{algorithm*}%
The main contribution of our work is a flare detection post-processing step for
HDBSCAN*. This section describes how the post-processing step works and
integrates with HDBSCAN* to form our FLASC algorithm (see
Algorithm~\ref{alg:fhdbscan}). FLASC starts by evaluating a flat HDBSCAN*
clustering, keeping track of the \emph{space tree} used in
HDBSCAN*~\citep{mcinnes2017hdbscan, mcinnes2017accelerated} to efficiently find
nearest neighbours. One noteworthy change from \cite{mcinnes2017hdbscan}'s
implementation is that we give all points the $0$-label when a single cluster is
allowed and selected and no \emph{cluster selection epsilon} is applied. This
enables FLASC to better analyse branching structures in data sets that contain a
single cluster. Then, for each selected cluster $C_j$, a branch detection step
is performed, explained in more detail below.

The concepts behind density-based clustering can also be applied to detect
branches within clusters by using an eccentricity measure in place of density,
as shown in Fig.~\ref{fig:eccentricity-clustering}. Peaks in an eccentricity
profile correspond to branches in the cluster, as shown in
Fig.~\ref{fig:eccentricity-clustering:profile} and
Fig.~\ref{fig:eccentricity-clustering:clusters}. We define eccentricity as the
distance to the cluster's centroid:
\begin{equation}
  \label{eq:eccentricity}
  e(\x_i) = d(\ctr_{C_j}, \x_i),
\end{equation}
where $\ctr_{C_j}$ is the cluster's membership-weighted average
(Fig.~\ref{fig:eccentricity-clustering:data}). This eccentricity measure can be
computed in $\mathcal{O}(N)$. Comparable to density contour clusters, an
\emph{eccentricity contour cluster} is a maximal, connected subset of points
with an eccentricity larger than some threshold $\{\x\;|\;e(\x) \ge e_t\}$. As
in functional persistence~\citep{carlsson2014flares}, eccentricity thresholds
filter out cluster cores, which separates branches and makes them detectable as
connected components. 

Similar to HDBSCAN*, we need a notion of connectivity between data points to
determine whether two points are part of the same eccentricity contour cluster
at a threshold $e_t$. In FLASC, we provide two solutions based on the cluster's
density scale in the form of \emph{cluster approximation graphs} $\G_k^{C_j}$:
the \emph{full} approximation graph and the \emph{core} approximation graph.
Both types contain a vertex for each point in the cluster $\x_i \in C_j$ but
differ in which edges they include. The \emph{full} approximation graph adds all
edges with $d_{mreach}(\x_i, \x_l) \le d_{max}^{C_j}$, where $d_{max}^{C_j}$ is
the largest distance in the cluster's minimum spanning tree (MST). The resulting
graph accurately describes the connectivity within the cluster at the density
where the last point joins the cluster. The \emph{space tree} constructed by
HDBSCAN* is used to retrieve these edges efficiently. The \emph{core}
approximation graph adds all edges with $d_{mreach}(\x_i, \x_j) \le
\max{\{\cd{(\x_i)}, \cd{(\x_j)}\}}$. The resulting graph accurately describes
the connectivity in the cluster's MST. Alternatively, this graph can be
considered as the cluster's subgraph from the $k$-nearest neighbour graph over
the entire data set. HDBSCAN* already extracted these edges when the core
distances were computed, so this approach has a lower additional cost. 

We can now recover the eccentricity contour tree as if it is were density
contour tree by applying HDBSCAN*'s clustering steps to the cluster
approximation graph with its edges weighted by $\min{\{e(\x_i), e(\x_l)\}}$.
This weighting ensures an edge has the eccentricity of the least eccentric point
it connects. Specifically, we adapt \cite{mcinnes2017accelerated}'s Union-Find
data structure to construct a single linkage dendrogram. The resulting hierarchy
is simplified using a minimum branch size $m_b$ to recover the condensed
branching hierarchy shown in Fig.~\ref{fig:eccentricity-clustering:hierarchy}.

\begin{figure*}[t!]
  \centering
  \subfigure[Probability average]{\includegraphics[width=.245\textwidth]{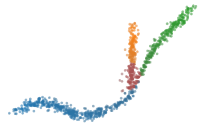}\label{fig:probability:average}}%
  \subfigure[Probability product]{\includegraphics[width=.245\textwidth]{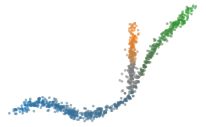}\label{fig:probability:product}}%
  \subfigure[Geodesic labels]{\includegraphics[width=.245\textwidth]{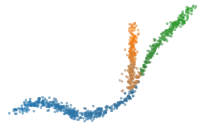}\label{fig:probability:max_membership}}%
  \subfigure[Geodesic membership]{\includegraphics[width=.245\textwidth]{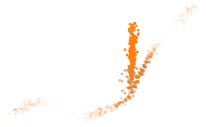}\label{fig:probability:weighted_membership}}%
  \caption{Different ways to combine cluster and branch membership
  probabilities. The cluster and branch probability average (a) and product (b)
  are visualised with desaturation. (c) Points labelled by the geodesically
  closest branch root---i.e., the point closest to the branch's weighted
  average---and desaturated as in (a). (d) Weighted branch membership for the
  orange branch is visualised by transparency. Branch memberships are computed
  from the traversal distance to the branch's root. }%
  \label{fig:probability}
\end{figure*}

HDBSCAN*'s EOM and leaf strategies are used to compute branch labels and
membership probabilities from these condensed hierarchies. Points that enter the
filtration after the selected branches have connected---i.e. points with the
noise label---are given a single non-noise label representing the cluster's
centre. Finally, the cluster and branch labels are combined. By default, points
in clusters with two or fewer branches are given a single label because two
branches are expected in all clusters, indicating the outsides growing towards
each other. The \emph{label sides as branches} parameter can be used to turn off
this behaviour and separate the ends of elongated clusters in the labelling. The
cluster and branch probabilities are combined by taking their average value
(Fig.~\ref{fig:probability:average}).

Other labelling and probability combinations are possible. For example, the
cluster and branch probability product more strongly emphasises the outsides of
the branches (Fig.~\ref{fig:probability:product}). As in
\cite{mcinnes2017hdbscan}, FLASC supports computing branch membership vectors
that describe how strongly a point $\x_i \in C_j$ belongs to each branch $B_b
\subset C_j$. These membership values are based on the geodesic distances in the
cluster approximation graph $\G_k^{C_j}$: $d_{geo}(\rt_{B_b}, \x_i)$, where
$\rt_{B_b}$ is the branch' root, i.e., the point closest to the branch's
membership-weighted average $\ctr_{B_b}$. The branch membership vectors can be
used to label central points by the closest branch root, as in
Fig.~\ref{fig:probability:max_membership}. Alternatively, a softmax function can
be used to convert $d_{geo}(\rt_{B_b}, \x_i)$ into the membership probabilities:
\begin{equation}%
  \label{eq:branch-membership}
  p(\x_i, B_{b}) = \frac{
    e^{c_b(\x_i, B_b) / t} 
  }{
    \sum_{B_{l} \in C_j}{e^{c_b(\x_i, B_l) / t}}
  },
\end{equation}
where $c_b(\x_i, B_b) = 1 / d_{geo}(\rt_{B_b}, \x_i)$ and $t$ is a temperature
parameter (Fig.~\ref{fig:probability:weighted_membership}).

Low persistent branches can be ignored using a \emph{branch selection
persistence} parameter, analogous to HDBSCAN*'s \emph{cluster selection
epsilon}~\citep{malzer2020hybrid}. As branches do not necessarily start at zero
centrality, \emph{branch selection persistence} describes the minimum
eccentricity range, rather than a single eccentricity threshold value. The
procedure that applies the threshold simplifies the condensed branch hierarchy
until all leaves have a persistence larger than the threshold.

\subsection*{Stability}%
\label{sec:fhdbscan:stability}
Stability is an important property of algorithms indicating that their output
differs only slightly when the input changes slightly. Two notions of stability
are relevant for FLASC: (1) the algorithm has to provide similar results when
run repeatedly on (different) samples of an underlying distribution, and (2) the
detected branch hierarchies have to represent the clusters' underlying topology
accurately. The deterministic density-based design of FLASC provides stability
in the first sense.

\cite{vandaele2021stable} analysed the second notion of stability for
graph-based branch detection, explaining that the graph approximation should
accurately represent the underlying shape and the graph-based centrality
function should accurately describe the points' centrality in a cluster's metric
space $(C_j, d_{mreach})$. For the \emph{normalised centrality} used
by~\cite{carlsson2014flares}, \cite{vandaele2021stable} show that the bound on
the bottleneck distances between true and empirical persistence diagrams is
tight if the metric distortion induced by the graph and its maximum edge weight
is small.

Both the \emph{full} and \emph{core} \emph{cluster approximation graph} used by
FLASC satisfy the low maximum edge weight requirement as their largest edge
weight is the minimum mutual reachability distance required for all points in
the cluster to be connected in the graph. Additionally, the metric distortion
should be small as only edges in the local neighbourhood of data points are
included because the clusters do not contain noise points.

The eccentricity function (Equation~\eqref{eq:eccentricity}) is more complex to
analyse. It is a $\mathrm{c}$-\emph{Lipschitz-continuous} function when
considered over the \emph{cluster centrality graph}'s edges: $$ |
\max{\{e(\x_i), e(\x_j)\}} - \max{\{e(\x_k), e(\x_l)\}} | \le \mathrm{c} \;
d_{mreach}(\x_i, \x_l), $$ where $\mathrm{c}$ is a constant describing the
continuity, $(\x_i, \x_j) \in \G_k^{C_j}$ and $(\x_k,   \x_l) \in \G_k^{C_j}$,
and the mutual reachability between $\x_i$ and $\x_l$ is the largest of the four
points. However, the position of the cluster's centroid $\ctr_{C_j}$ influences
how well the function represents the centrality in the cluster's metric space
$(C_j, d_{mreach})$. We aim to show the current approach strikes a good balance
between computational cost and stability in the experiments presented in the
next section.

\subsection*{Computational Complexity}%
\label{sec:fhdbscan:complexity}
The algorithm's most computationally expensive steps are constructing the
\emph{full cluster approximation graphs} and computing their single linkage
hierarchies. Naively, the worst-case complexity for creating a \emph{cluster
approximation graph} is $\mathcal{O}(n_c^2)$, where $n_c$ is the number of
points in the cluster. Usually, the average case is much better because the
approximation graphs rarely are fully connected. After all, HDBSCAN*'s noise
classification limits the density range within the clusters. Furthermore, the
\emph{space tree} that we re-use from the HDBSCAN* clustering step provides fast
asymptotic performance for finding the graph's edges. The exact run-time bounds
depend on the data properties. They are challenging to describe (as explained
in~\cite{mcinnes2017accelerated}), but an average complexity proportional to
$\mathcal{O}(n_e\log{N})$ is expected, where $n_e$ is the number of edges in the
approximation graph. Computing single linkage hierarchies from the \emph{cluster
centrality graphs} is possible in $\mathcal{O}(n_e \alpha(n_e))$
using~\cite{mcinnes2017hdbscan}'s Union-Find implementation adapted to ignore
edges between data points that are already in the same connected component
($\alpha$ is the inverse Ackermann function).
Like \cite{mcinnes2017accelerated}, we feel confident that FLASC achieves
sub-quadratic complexity on average, which we demonstrate in a practical example
in the next section.

\section*{Experiments}%
\label{sec:experiments}
Stability and computational performance are essential properties for clustering
algorithms to be useful in practice. This section presents two synthetic
benchmarks that compare these properties between FLASC's, HDBSCAN*, and kMeans.
In addition, we demonstrate two exploration use cases where detecting
branch-based subgroups and branch hierarchies helps to understand the data's
structure.

\subsection*{FLASC Stability}%
\label{sec:experiments:stability}
This first synthetic benchmark compares the ability of FLASC, HDBSCAN*, and
kMeans to detect branch-based subgroups that do not contain a density maximum.
The benchmark is designed to answer the following research question:
\begin{quote}
  How does FLASC's branch detection ability compare to kMeans and HDBSCAN* in
  terms of accuracy and stability?
\end{quote}
kMeans was selected for this comparison because it is a popular, fast clustering
algorithm that should be able to detect branch-based subgroups when given the
correct number of subgroups a priori. HDBSCAN* was included to show that it
cannot reliably find subgroups that do not contain a density maximum,
demonstrating that our branch-detection post-processing step enables the
algorithm to find an additional type of pattern. 

We expect both kMeans and FLASC to be able to detect branching structures, while
HDBSCAN* will struggle due to the lack of density maxima in the to-be-detected
subgroups. Furthermore, we expect FLASC to have a higher stability than
kMeans---in terms of difference between predictions on multiple samples of the
same underlying distribution---as it does not rely on converging from a random
initialisation. The next two subsections explain how the data for this benchmark
was generated and how we measured the algorithm's accuracy and stability,
respectively.

\subsubsection*{Datasets}%
\label{sec:experiments:stability:data}
Two-dimensional data sets containing a single cluster with $3$ branches laid out
as a three-pointed star were generated
(Fig.~\ref{fig:experiments:stability:centroids}). The branches span from the
centre outwards and are spread out equally across the $2$D plane---i.e., the
angle between adjacent branches is $120$ degrees. Each branch has a length $b_l$
and contains $100$ points exponentially spaced from the inside out.
Consequently, the density is highest at the cluster's centre and lowest in the
branch ends, and this difference increases with the branch length. Normally
distributed noise ($\mu=0$) is added to the points' coordinates using a noise
ratio parameter $n_r$ to determine the distribution's standard deviation
$\sigma$:
\begin{equation}%
  \label{eq:experiments:stability:noise-ratio}%
  n_r = \frac{4\sigma}{\sqrt{\frac{1}{2}} b_l}.
\end{equation}
Here, $n_r$ indicates that approximately $95\%$ of the sampled noise values fall
within $\pm \sqrt{\frac{1}{2}} n_r b_l$ of zero. Consequently, approximately
$90\%$ of points are moved less than $n_r b_l$ from their original 2D position.

\subsubsection*{Evaluation and settings}%
\label{sec:experiments:stability:evaluation}
\begin{figure*}[t!]
  \centering
  \includegraphics[width=1\textwidth]{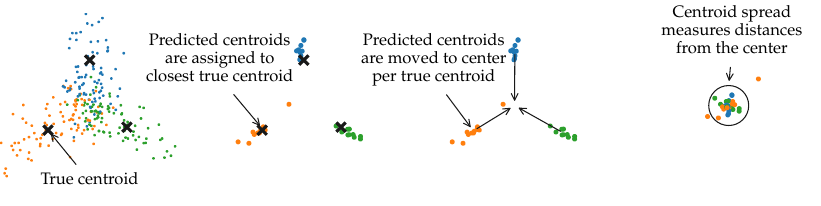}
  \caption{Explanatory figure for the centroid spread stability measure.
  Predicted centroids---weighted average coordinates---are computed for each
  predicted subgroup and assigned to the closest ground-truth centroid. The 95
  percentile centroid spread can be interpreted by translating each group's
  average to the origin and finding the radius for a circle that includes 95\%
  of the predicted centroids.}%
  \label{fig:experiments:stability:centroids}%
\end{figure*}%
The data sets were sampled with varying branch lengths $b_l$ ($2$ to $100$ in
$10$ exponentially spaced steps) and noise ratios $n_r$ ($0$ to $1$ in $10$
exponentially spaced steps). Ten data sets were sampled for each parameter
combination, resulting in $1000$ point clouds. FLASC was tuned to find a single
cluster by enabling the \emph{allow single cluster} parameter and setting
\emph{min samples} $k=5$ and \emph{min cluster size} $m_c=25$. Its branch
detection parameters were optimised in a grid search: using the \emph{full} and
\emph{core} approximation graphs, both branch selection strategies, and varying
the \emph{min branch size} $m_b$ between $2$ to $24$ in steps of $2$. Labels for
central points were set to the geodesically closest branch root (as in
Fig.~\ref{fig:probability:max_membership}). HDBSCAN* was tuned to find multiple
clusters using \emph{min samples} $k=5$. Its other parameters were optimised in
a grid search: varying \emph{min cluster size} $m_c=$ between $2$ to $24$ in
steps of $2$, and using both cluster selection strategies. For, kMeans, we set
$k=3$, using our prior knowledge that there are three branch-based subgroups in
the data.

For each evaluation, the resulting data point labels and probabilities were
stored and used to computed the Adjusted Rand Index
(ARI)~\citep{hubert1985comparing}. ARI values describe the agreement between
ground truth and assigned labels adjusted for chance. In addition, the
algorithm's stability was measured using the spread of the predicted branch
centroids (see, Fig.~\ref{fig:experiments:stability:centroids}). Here, the
weighted average coordinates per detected subgroup were used as \emph{predicted
centroid}. These predicted centroids were assigned to the closest ground-truth
centroid. Their spread was quantified as the distance between the predicted
centroids and their unweighted average per ground-truth centroid. This value was
normalised by the branch length $b_l$ to allow comparisons across branch
lengths.

Finally, we selected the parameter values that maximised the average ARI and
minimised the average centroid spread over all data sets. For FLASC, these
parameter values were: \emph{min branch size} $m_b = 12$, the core cluster
approximation graph, and leaf branch selection strategy. Larger values for $m_b$
provided similar performance values, indicating the algorithm is not very
sensitive to $m_b$ as long as its value is large enough to exclude small noisy
structures. HDBSCAN* performed best with \emph{min cluster size} $m_c=8$ and the
leaf cluster selection strategy.

\subsubsection*{Results}%
\label{sec:experiments:stability:results}
\begin{figure*}[t!]
  \centering
  \subfigure[True and predicted subgroups]{\includegraphics[width=1\textwidth]{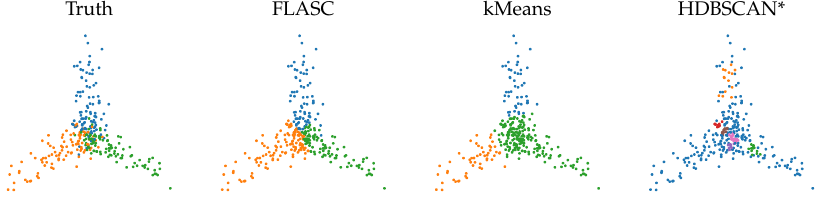}\label{fig:experiments:stability:labelling}}
  \subfigure[Average ARI]{\includegraphics[width=0.5\textwidth]{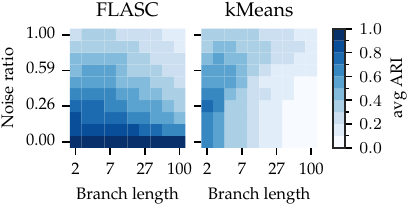}\label{fig:experiments:stability:ari}}%
  \subfigure[95\% centroid spread]{\includegraphics[width=0.5\textwidth]{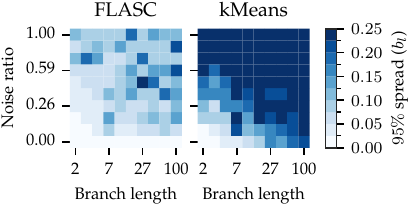}\label{fig:experiments:stability:spread}}
  \caption{Results for the stability benchmark. (a) One point cloud with  $b_l =
  18$ and $n_r = 0.47$ coloured by ground truth and predicted labels that
  characterise the algorithms' behaviours. (b) Heatmap with average ARI values
  for FLASC and kMeans over all branch lengths and noise ratios. (c) Heatmap
  with 95\% centroid spread values for FLASC and kMeans over all branch lengths
  and noise ratios.}%
  \label{fig:experiments:stability-results}%
\end{figure*}%
Figure~\ref{fig:experiments:stability:labelling} demonstrates which subgroups
the algorithms detect on a point cloud with $b_l = 18$ and $n_r = 0.47$. The
true labels indicate that most confusion is expected in the dense centres, as
the true labels are not spatially separable in those locations. Furthermore,
FLASC neatly separated central points by the closest branch. kMeans, on the
other hand, generally assigned all central points to one branch and limits the
other branches to non-central points. Finally, as expected, HDBSCAN* did not
detect the branch-based subgroups.

The patterns identified in the previous paragraph generalised to other branch
lengths and noise ratios, as shown by the average ARI in
Fig.~\ref{fig:experiments:stability:ari}. FLASC achieved ARI values $\ge 0.8$ on
all branch lengths for low noise ratios. Its performance degrades as more noise
is introduced, where the decrease in ARI is stronger for large branch lengths.
This pattern can be explained by the confusion on central points, as there are
more central points with longer branch lengths and larger noise ratios. kMeans
achieved ARI values $\le 0.8$. Its performance decreased most strongly with the
branch length, which can be explained by the assignment of all central points to
one branch.

The 95 percentile centroid spread, shown in
Fig.~\ref{fig:experiments:stability:spread}, indicates FLASC provided more
stable results than kMeans. Both algorithms behaved in a similar pattern: the
spread was lowest with low noise values and small branch lengths. When either
parameter increased, the spread increased as well. However, FLASC achieved lower
spread values in these cases, mostly remaining below $0.25b_l$. For kMeans, the
spread was larger than $0.25b_l$ in the upper $b_l$--$n_r$ triangle. The
behaviour on central points explains these patterns.

All in all, this benchmark showed that both FLASC and kMeans are able to detect
the branch based subgroups. However, FLASC more accurately segmented the central
points, which resulted in a higher accuracy and stability. In addition, FLASC
achieved this result without being given the correct number of subgroups a
priori.

\subsection*{Computational Performance}%
\label{sec:experiments:performance}
Now we turn to the computational performance. This second synthetic benchmark is
designed to answer the following research question:
\begin{quote}
  How does FLASC's compute cost compare to HDBSCAN*, kMeans, and fastcluster?
\end{quote}
kMeans is included in this comparison because it is a fast, popular clustering
algorithm. HDBSCAN* is included to determine how much extra cost FLASC
introduces. Fastcluster is included as an example of a quadratically scaling
implementation (as shown in~\cite{mcinnes2017accelerated}). 

Given the challenges in accurately benchmarking the computational performance of
algorithms~\citep{kriegel2017black}, we limit this comparison to the trends in
run time scaling over data set size and number of dimensions for specific
implementations.

\subsubsection*{Datasets}%
\label{sec:experiments:performance:data}
A Gaussian random walk process was used to generate data sets that contain
non-trivially varying densities and branching structures in a controlled
environment. For a space with $d$ dimensions, $c$ uniform random starting points
were sampled in a volume that fits five times the number of to-be-generated
clusters. Then, five $50$-step random walks were sampled from each starting
point. Every step moved along one of the dimensions with a length sampled from a
normal distribution ($\mu=0$, $\sigma=0.1$). The resulting point clouds have
more varied properties than the Gaussian blobs used
in~\cite{mcinnes2017accelerated}'s run time comparison of HDBSCAN*. Note that
the number of (density-based) clusters in each point cloud may differ from the
number of starting points $c$ due to possible overlaps or sparse regions in the
random walks.

\subsubsection*{Evaluation and settings}%
\label{sec:experiments:performance:evaluation}
The random walk data sets were generated with varying numbers of dimensions
($2$, $8$, $16$) and starting points ($2$ to $800$ in $10$ exponentially spaced
steps). Ten data sets were sampled for each combination of parameters, resulting
in a total of $300$ point clouds.

The algorithms were compared using their Python implementations: fastcluster
version 1.1.26~\citep{mullner2013fastcluster}, kMeans version
1.3.2~\citep{pedregosa2011scikit}, HDBSCAN* version
0.8.28~\citep{mcinnes2017hdbscan}, and FLASC version 0.1.3 with the \emph{full}
and \emph{core} approximation graphs. HDBSCAN* and FLASC were evaluated with
\emph{min samples} $k=10$, \emph{min cluster size} $m_c=100$, \emph{min branch
size} $m_b=20$, \emph{allow single cluster} enabled, and their multiprocessing
support disabled to better describe the algorithms' intrinsic complexity. These
parameter values allow the algorithms to find clusters and branches that are
slightly smaller than how they are generated. Fastcluster's default parameter
values were used, resulting in a single linkage dendrogram. kMeans was evaluated
with $k=c$, effectively attempting to recover one cluster for each starting
point.

Time measurements were performed with a $5.4$GHz AMD R7 7700X processor. Each
algorithm was evaluated on each data set once, recording the run time and number
of detected clusters. The smallest data set for which an algorithm required more
than $80$ seconds was recorded for each number of dimensions. Larger data sets
were not evaluated for those algorithms.

\begin{figure*}[t!]
  \centering
  \subfigure[2D]{\includegraphics[width=.32\textwidth]{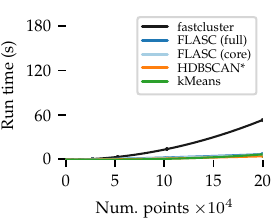}%
  \label{fig:experiments:performance:2}}%
  \subfigure[8D]{\includegraphics[width=.32\textwidth]{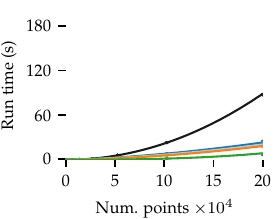}%
  \label{fig:experiments:performance:8}}%
  \subfigure[16D]{\includegraphics[width=.32\textwidth]{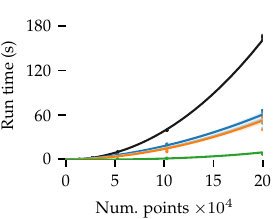}%
  \label{fig:experiments:performance:16}}%
  \caption{Benchmark run times (s) over the data set size and number of
  dimensions. The algorithms' scaling behaviours are visualised by quadratic
  regression lines relating compute time to the number of points in log--log
  space, shown in linear space. The shaded areas around each line indicate the
  regression's $95\%$ confidence interval.}
  \label{fig:flasc:experiments:performance}
\end{figure*}
  
\subsubsection*{Results}%
\label{sec:experiments:performance:results}
Figure~\ref{fig:flasc:experiments:performance} shows run times in seconds over
the data set size and number of dimensions. There are three patterns of note.
Firstly, fastcluster shows the steepest quadratic trend, resulting in the
longest run times. FLASC and HDBSCAN* show shallower trends on the $2$ and $8$
dimensional data sets but approach fastcluster's quadratic trend in the $16$
dimensional case. This is expected because querying \emph{space trees} becomes
more expensive with more dimensions~\citep{weber1998quantitative}. Secondly,
HDBSCAN* and both FLASC variants scale similarly in all three conditions.
HDBSCAN* is consistently the fastest of the three, followed by FLASC with the
\emph{core} approximation graph. However, their run time differences diminish in
the higher dimensional cases, indicating that the additional cost of branch
detection is relatively low compared to the cost of detecting the clusters.
Finally, kMeans had the shallowest scaling trend in all three conditions,
indicating that it is the fastest algorithm of the four.

In conclusion, both FLASC variants' computational performance scale similarly to
HDBSCAN* and the more dimensions the data contains, the smaller the scaling
trend differences between the algorithms. 

\subsection*{Use Case: Diabetes Types}%
\label{sec:experiments:cases:diabetes}
Next, we present a data exploration case in which identifying branch-based
subgroups is essential to understand the data. \cite{reaven1979diabetes}
attempted to clarify a ``horse shoe''-shaped relation between glucose levels and
insulin responses in diabetes patients. Three of the metabolic variables they
measured turned out to be very informative in a 3D scatterplot, showing a dense
core with two less-dense branches, which they considered unlikely to be a single
population. Seeing that plot was instrumental in their understanding of the
data~\citep{miller1985discussion}.

More recently, \cite{singh2007mapper} used the same data set to demonstrate how
Mapper with a density-based lens function visualises these flares without
manually specifying which dimensions to plot. Their analysis leverages the
flares' lower density, allowing them to be detected without a centrality metric.
In general, though, local density minima do not always relate to branches,
especially for data sets with multiple branching clusters.

In this use case, we show how FLASC can detect the branching pattern in this
data set and classify the observations by their branch without manually
extracting the flares from a visualisation.

\subsubsection*{Evaluation and settings}%
\label{sec:experiments:cases:diabetes:settings}
The data set---obtained from~\cite{andrews1985diabetes}---contains five
variables describing 145 subjects: the relative weight, the plasma glucose level
after a period of fasting, the steady-state plasma glucose response (SSPG), and
two areas under a curve---one for glucose (AUGC) and one for insulin
(AUIC)---representing the total amount observed during the experimental
procedure described in~\cite{reaven1979diabetes}. All five variables were
z-score normalised and used to compute the Euclidean distance between subjects.

Both FLASC and HDBSCAN* were evaluated on the normalised data set. FLASC was
tuned to find a single cluster with multiple branches by setting \emph{min
samples} $k=5$, \emph{min cluster size} $m_c=100$, \emph{min branch size}
$m_b=5$, and enabling \emph{allow single cluster}. HDBSCAN* was tuned to find
multiple clusters with \emph{min samples} $k=5$ and \emph{min cluster size}
$m_c=10$.

\subsubsection*{Results}%
\label{sec:experiments:cases:diabetes:results}
\begin{figure*}
  \centering
  \subfigure[FLASC]{%
    \includegraphics[width=.32\textwidth, trim={0 1cm 0
    1cm},clip]{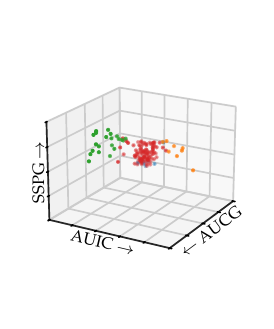}%
    \label{fig:experiments:cases:diabetes:flasc}}%
  \subfigure[HDBSCAN*]{%
    \includegraphics[width=.32\textwidth, trim={0 1cm 0
    1cm},clip]{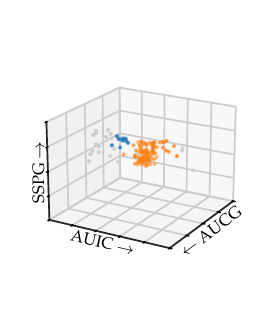}%
    \label{fig:experiments:cases:diabetes:hdbscan}}%
  \caption{Subgroups detected by (a) FLASC and (b) HDBSCAN* shown in a 3D
    scatterplot over the area under the plasma glucose curve (AUGC), the area
    under the insulin curve (AUIC), and the steady state plasma glucose response
    (SSPG) from~\cite{reaven1979diabetes}. Grey points were classified as
    noise.}%
  \label{fig:experiments:cases:diabetes}%
\end{figure*}
Figure~\ref{fig:experiments:cases:diabetes} shows the detected subgroups encoded
using colour on the 3D scatterplot. FLASC's classification
(Fig.~\ref{fig:experiments:cases:diabetes:flasc}) distinguishes the branches
from the central core. The algorithm also finds a low-persistent flare
representing the central core's bottom. This flare could be ignored by
specifying a persistence threshold. In contrast, HDBSCAN*'s classification
(Fig.~\ref{fig:experiments:cases:diabetes:hdbscan}) does not find the branches.
Instead, it finds part of the left branch as a small low-persistent cluster and
merges most of the right branch with the central core.

All in all, this case study demonstrated how FLASC detects branch-based
subgroups that do not contain local density maxima without having to specify the
relevant features in advance or extract the subgroups visually. Practically,
FLASC would have made it easier for researchers to detect the three groups in
this data set, which was relevant for understanding diabetes and its causes.

\subsection*{Use Case: Cell Development}%
\label{sec:experiments:cases:elegans}
Finally, we demonstrate a use case where detecting a branch hierarchy is
important for understanding the data set. Specifically, we analyse
\cite{packer2019elegans}'s cell development atlas for the \emph{C. Elegans}, a
small roundworm that is often used in biological studies. They analysed gene
expressions in \emph{C. Elegans} embryos to uncover the trajectories along which
cells develop. Broadly speaking, this data set describes what happens in cells
as they develop from a single egg cell into all the different tissues that exist
within fully grown \emph{C. Elegans} worms.

After pre-processing, the data set appears to contain both cluster and branching
structures when viewed in \cite{packer2019elegans}'s 3D projection. In this use
case, we demonstrate that FLASC's branch hierarchy provides interesting
information about the data set's structure even when the main subgroups can be
detected as clusters.

\subsubsection*{Evaluation and settings}%
\label{sec:experiments:cases:elegans:methods}
The data and pre-processing scripts were obtained from
Monocle~3's~\citep{cao2019monocle3} documentation~\citep{monocle3documentation}.
The pre-processing stages normalise the data, extract the 50 strongest PCA
components, and correct for batch effects using algorithms
from~\cite{haghverdi2018batcheffects}. HDBSCAN* was evaluated on the
pre-processed data with an angular distance metric because the cosine distance
metric is not supported in its optimised code path. HDBSCAN* was tuned to find
multiple clusters with \emph{min samples} $k=5$ and \emph{min cluster size}
$m_c=50$. FLASC was evaluated on a 3D UMAP~\citep{mcinnes2018umap} projection
that denoised the data set. Using three instead of two dimensions reduces the
chance of branch overlaps in the embedding. UMAP used the angular distance
metric to find $30$ nearest neighbours. The \emph{disconnection distance}
parameter was set to exclude the 16-percentile least dense points detected by
HDBSCAN*, thereby preventing shortcuts across the data's structure. The largest
connected component in the resulting UMAP graph was projected to 3D while
varying the repulsion strength to avoid crossings and ensure connected
structures remained close. The same layout procedure was used to project the
graph to 2D to visualise the data. FLASC was tuned to detect the branching
hierarchy of the dataset as a single cluster by selecting \emph{min samples}
$k=3$, \emph{min cluster size} $m_c=500$, \emph{min branch size} $m_b=50$,
enabling \emph{allow single cluster}. Branches were detected using the
\emph{core} cluster approximation graph and selected with the leaf strategy.

\subsubsection*{Results}%
\label{sec:experiments:cases:elegans:results}
\begin{figure*}[t!]
  \centering
  \subfigure[HDBSCAN*]{%
    \includegraphics[width=.48\textwidth]{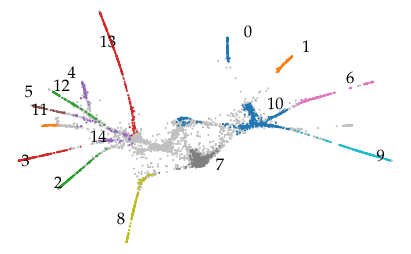}%
    \label{fig:experiments:cases:elegans:clusters}}%
  \subfigure[FLASC]{%
    \includegraphics[width=.48\textwidth]{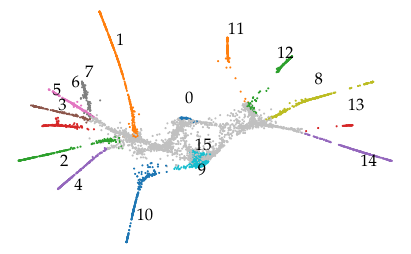}%
    \label{fig:experiments:cases:elegans:branches}} \subfigure[Cluster
  hierarchy]{%
    \includegraphics[width=.48\textwidth]{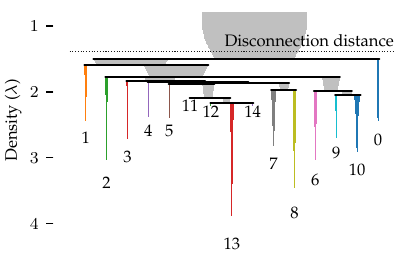}%
    \label{fig:experiments:cases:elegans:cluster_tree}}%
  \subfigure[Branch hierarchy]{%
    \includegraphics[width=.48\textwidth]{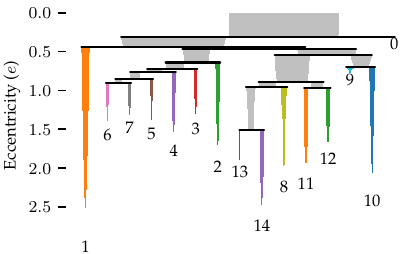}%
    \label{fig:experiments:cases:elegans:branch_tree}} \caption{Results for the
  single cell gene expression use case using 50 dimensional pre-processed data
  from~\cite{packer2019elegans}. (a) 2D UMAP projection~\citep{mcinnes2018umap}
  coloured by HDBSCAN* clusters detected in the pre-processed data. (b) The same
  projection coloured by FLASC branches detected from a 3D UMAP projection. (c)
  and (d) show cluster and branch hierarchies, respectively. The icicle plots
  were adapted from~\cite{mcinnes2017hdbscan} to indicate selected clusters with
  colour and labels.}%
  \label{fig:experiments:cases:elegans:projections}%
\end{figure*}
Figures~\ref{fig:experiments:cases:elegans:branches}
and~\ref{fig:experiments:cases:elegans:clusters} show the 2D projections. Data
points are coloured to indicate the detected clusters and branches,
respectively. There are two main differences between the branches and the
clusters. Firstly, two regions where branches merge are detected as clusters,
namely clusters 10 and 14. These regions do not have a distinct branch label,
but are identifiable in the branch hierarchy
(Fig.~\ref{fig:experiments:cases:elegans:branch_tree}). Secondly, all but one
branch---branch 13---are also detected as a cluster in this dataset, indicating
the branches represent regions with high local density. Considering that the
branches correspond to developmental end-states, it is unsurprising that local
density maxima occur within them. One could imagine that the variation in gene
expression is higher during development and that fully developed cells are
observed more frequently. Both scenarios could cause these local density maxima.

More interesting are the differences between the cluster and branch hierarchies.
Figures~\ref{fig:experiments:cases:elegans:cluster_tree}
and~\ref{fig:experiments:cases:elegans:branch_tree} visualise these hierarchies
as an icicle plot. These designs are adapted from~\cite{mcinnes2017hdbscan} to
indicate the selected branches and clusters using colour and a label. Segment
widths encode the number of points in the tree below the segment. The
hierarchies highlight that while HDBSCAN* detects the branches as clusters, it
does not capture the trajectories. For example, the hierarchy does not reflect
that clusters 0 and 1 connect to the whole data set through cluster 10. FLASC's
branch hierarchy, on the other hand, more closely resembles the data's shape.
For example, the hierarchy describes the embedding's left and right side with 5
and 6 branches, respectively. In general, branches that merge into the cluster
near each other are also close in the branch condensed tree. For branches
connected to multiple other branches in the cluster approximation graph, only
the most eccentric connection is captured by the branch-condensed tree. 

All in all, this use case demonstrated that FLASC's branch hierarchy provides
information about the data's shape that may not be obvious from cluster
hierarchies. In addition, we found it beneficial to suppress noisy connectivity
using dimensionality reduction techniques when detecting branches.

\section*{Discussion}%
\label{sec:discussion}
Two synthetic benchmarks and two real-world use cases were performed to
demonstrate FLASC and its properties. We start our discussion by providing
remarks for each benchmark and use case in order.

The stability benchmark compared FLASC and kMeans in their ability to detect
branches that do not contain a density maximum. The benchmark quantified
performance in terms of ARI and a centroid spread measure that describes output
similarity between multiple samples of the same underlying distribution. While
both algorithms were able to detect the branches, FLASC performed better than
kMeans on both measures. In addition, kMeans requires specifying the correct
number of subgroups a priori, which is difficult with unfamiliar data. This
analysis could be expanded to investigate how well FLASC deals with unequal
branch lengths. The weighted average data point---and eccentricity measure as a
result---may not accurately reflect the centre of such clusters. Consequently,
FLASC's branch hierarchy will be a less accurate representation of the
underlying topology but should still detect the branches.
Monocle~3~\citep{qiu2017graphembedding} deals with this problem by selecting the
centre point in a projection manually~\citep{monocle3documentation}. Other
eccentricity measures discussed below could also improve FLASC's performance in
such cases.

The computational performance benchmark demonstrated that FLASC's computational
performance scales similar to HDBSCAN*. The scaling trends appear to become more
similar as the data contained more dimensions. Neither algorithm was evaluated
with multiprocessing enabled, which can introduce run-time differences in
practical applications. In addition, extracting the \emph{full} cluster
approximation graph can be more expensive than reported, depending on the data's
characteristics. kMeans provided even quicker run times, but is limited in  
usability due to lower stability and its predefined number of clusters.

The diabetes types use case demonstrated a real-world dataset in which branches
that are not detectable as density-based clusters represent meaningful
subgroups. FLASC is designed to detect such branches without knowing they exist
a priori or extracting them from a visualisation manually. The cell development
use case showed how FLASC behaves on a more complex data set. Here, the
subgroups were detectable by both FLASC and HDBSCAN*. FLASC still provides a
benefit for exploration because its branch hierarchy more closely resembles the
data's shape. Structure learning algorithms, as described in Section ``Related
Work -- Structure Learning'', can provide even more information about the data's
shape at more computational costs.

\subsection*{FLASC's practical value}%
\label{sec:discussion:why}
As demonstrated by the cell development use case, the argument that branches are
not detectable as clusters only applies when they do not contain local density
maxima. Subpopulations tend to have some location in feature space where
observations are more likely. These locations are detectable as local density
maxima, allowing data points surrounding them to be classified as a particular
cluster. If one is only interested in the existence of subgroups, then FLASC
only provides a benefit on datasets where relevant subgroups are sparse (e.g.,
the diabetes types use case). If one is also interested in the clusters' shapes,
then FLASC's branch hierarchy provides information that cannot be extracted from
a cluster hierarchy. We envision FLASC as a valuable tool for exploring
unfamiliar data, providing guidance into which subpopulations exist and
informing follow-up questions. Knowing that a cluster may represent multiple
subgroups can be very relevant.

\subsection*{Alternative eccentricity metrics}%
\label{sec:discussion:eccentricity}
The presented FLASC algorithm uses a geometric distance-to-centroid metric to
describe how eccentric data points are within a cluster
(Fig.~\ref{fig:eccentricity-clustering:data}, Eq.~\ref{eq:eccentricity}). An
interesting alternative is an unweighted geodesic eccentricity, which measures
path-lengths between each data point and the cluster's root point in the cluster
approximation graph. Here, the root point can be chosen as the data point
closest to the cluster's centroid, as we did for the branch-membership vectors
(Fig.~\ref{fig:probability}). Such a geodesic eccentricity would agree with the
notion that distances in high dimensional data may not accurately reflect
distances along the intrinsic structure of a data set, which was one of the
motivations for Reversed Graph Embeddings~\citep{mao2017graphembedding}. It
would also be closer to the maximum shortest-path centrality metric used
by~\citep{vandaele2021stable}.

Several trade-offs between the geometric and geodesic eccentricity metrics made
us choose the geometric one:
\begin{itemize}
  \item Computing the geodesic eccentricity is more expensive because it
  requires an additional traversal over the entire cluster approximation graph.
  The extra cost, however, should be low compared to other parts of the
  algorithm.
  \item The resolution of the unweighted geodesic eccentricity is lower, as it
  expresses the number of edges to the root point. As a result, zero-persistent
  branches are more likely to occur. In addition, it reduces the detectability
  of small branches that are well connected. On the other hand, that can be seen
  as beneficial noise suppression. In addition, the \emph{branch selection
  persistence} parameter becomes more interpretable and would represent the
  traversal depth of a branch in the approximation graph.
  \item The cluster's centroid may lie outside of the cluster itself, resulting
  in a root point and eccentricity values that do not accurately describe its
  centre. For example, imagine a U-shaped cluster. The centroid would lie in
  between the two arms, and the root would lie in one of the arms. As a result,
  the geodesic metric would find one smaller and one larger branch rather than
  two equal branches. On the other hand, the geometric eccentricity finds the
  two branches and the connecting bend as three separate groups. Confusingly, it
  also contains two regions with a local eccentricity maximum, which FLASC gives
  a single label. Placing the root at an arbitrary eccentric location as
  in~\cite{ge2011reebskeleton} avoids this issue but necessitates a different
  interpretation of the branch hierarchy and branch probability.   
\end{itemize}

FLASC's general process can also be used with metrics that capture other aspects
than eccentricity. At its core, FLASC consists of two filtrations, one to
determine the connectivity between data points and one to analyse a signal on
the resulting graph. The process would then describe how many distinct local
minima (or maxima) of the metric exist within the clusters. The resulting
interpretation does not have to relate to the cluster's shape.

One could even interpret FLASC as two applications of HDBSCAN*: one over the
density and one over the eccentricity. This perspective raises a possible
improvement to the algorithm by translating the mutual reachability concept to
the centrality metric. The idea of `pushing away points in low-density regions'
can also be applied to the centrality and would emphasise the centrality
difference between the centre and branch ends. Additionally, smoothing the
centrality profile by incorporating neighbouring values could improve the
algorithm's robustness to noise. The additional computational cost should be
low, as points' neighbours are already known when the centrality is computed.
Another way to improve noise robustness could be to implement the \emph{mutual}
$k$-nearest neighbour approach used by~\cite{dalmia2021mutual} to improve
UMAP projections. It would provide a subgraph of the \emph{core} approximation
graph that better reflects the cluster's connectivity in high dimensional data
sets. We leave evaluating these ideas for future work.

\subsection*{Visually summarising data's shape}%
\label{sec:discussion:visualisations}
A strength of Mapper~\citep{singh2007mapper} and Reversed Graph
Embeddings~\citep{mao2017graphembedding} is that they can summarise the data's
shape using intuitive visualisations. While FLASC's branch-condensed tree
provides some information about the clusters' shapes, interpreting the shape is
not trivial. Studying how well two-dimensional layouts of FLASC's cluster
approximation graphs work as shape summarising visualisations would be an
interesting future research direction. These graphs directly encode the
connectivity used by the algorithm. Another benefit is that---unlike in
Mapper---all (non-noise) data points are represented in the graphs once.
Directly visualising the graphs, however, probably does not scale to larger
sizes in terms of computational cost for the layout algorithm and visual
interpretability. Ways to summarise the networks would have to be found, which
could be based on kMeans centroids like in Reversed Graph
Embeddings~\citep{mao2017graphembedding}, local density maxima in the cluster,
or a Reeb-Graph approach similar to~\cite{ge2011reebskeleton}.

\section*{Conclusion}
\label{sec:conclusion}
We presented the FLASC algorithm that combines HDBSCAN* clustering with
branch-detection post-processing step. We have shown that the algorithm can
detect branch-based subgroups that do not contain local density maxima in
real-world data without specifying features of interest or extracting the
branches from a visualisation manually. In addition, we demonstrated that
branching hierarchies found by FLASC can provide information about the data's
shape that is not present in HDBSCAN*'s cluster hierarchy. Two synthetic
benchmarks demonstrated FLASC's stability and indicated FLASC's computational
performance scales similarly to HDBSCAN*.

\section*{Acknowledgements}
We thank Kris Luyten for his comments on an early version of the manuscript.

\section*{Data Availability}
The data and code used in this paper are available at
\url{https://zenodo.org/records/13326252}.

\section*{Funding Statement}
The work of D.M. Bot was supported by Hasselt University BOF grant [BOF20OWB33].
The work of J. Peeters was supported by Hasselt University BOF grant
[BOF21DOC19]. The work of J. Aerts was supported by KU Leuven grant STG/23/040.

\bibliography{references}
\end{multicols}
\end{document}